\documentclass[sigconf]{acmart}
\copyrightyear{2023}
\acmYear{2023}
\setcopyright{acmlicensed}
\acmConference[MM '23] {Proceedings of the 31st ACM International Conference on Multimedia}{October 29--November 3, 2023}{Ottawa, ON, Canada.}
\acmBooktitle{Proceedings of the 31st ACM International Conference on Multimedia (MM '23), October 29--November 3, 2023, Ottawa, ON, Canada}
\acmPrice{15.00}
\acmDOI{10.1145/3581783.3611888}
\acmISBN{979-8-4007-0108-5/23/10}

\AtBeginDocument{%
  }

\acmSubmissionID{887}

\begin{document}

\title{Zero-shot Skeleton-based Action Recognition via Mutual
Information Estimation and Maximization}

\author{Yujie Zhou}
\affiliation{%
  \institution{Gaoling School of Artificial Intelligence, Renmin University of China}
  \streetaddress{59 Zhong Guan Cun Avenue}
  \city{Beijing}
  \country{China}}
\email{yujiezhou@ruc.edu.cn}

\author{Wenwen Qiang}
\affiliation{%
 \institution{University of Chinese Academy of Sciences}
 \institution{Institute of Software Chinese Academy of Sciences}
  \city{Beijing}
  \country{China}}
\email{wenwen2018@iscas.ac.cn}

\author{Anyi Rao}
\affiliation{%
  \institution{Stanford University}
  \streetaddress{353 Jane Stanford Way}
  \city{Stanford, CA}
  \country{USA}}
\email{anyirao@stanford.edu}

\author{Ning Lin}
\affiliation{%
  \institution{Gaoling School of Artificial Intelligence, Renmin University of China}
  \streetaddress{59 Zhong Guan Cun Avenue}
  \city{Beijing}
  \country{China}}
\email{linning51400@ruc.edu.cn}

\author{Bing Su}
\authornote{Corresponding author}
\affiliation{%
  \institution{Gaoling School of Artificial Intelligence, Renmin University of China}
  \institution{Beijing Key Laboratory of Big Data Management and Analysis Methods}
  \city{Beijing}
  \country{China}}
\email{subingats@gmail.com}

\author{Jiaqi Wang}
\affiliation{%
  \institution{Shanghai AI Laboratory}
  \city{Shanghai}
  \country{China}}
\email{wjqdev@gmail.com}

\renewcommand{\shortauthors}{Yujie Zhou, Wenwen Qiang, Anyi Rao, Ning Lin, Bing Su, Jiaqi Wang}

\begin{abstract}
Zero-shot skeleton-based action recognition aims to recognize actions
of unseen categories after training on data of seen categories. 
The key is to build the connection between visual and semantic space 
from seen to unseen classes. 
Previous studies have primarily focused on encoding sequences into a singular feature vector,
with subsequent mapping the features to an identical anchor point within the embedded space.
Their performance is hindered by 1) the ignorance of the global visual/semantic distribution alignment,
which results in a limitation to capture the true interdependence between the two spaces.
 2) the negligence of temporal information since the frame-wise features with rich action clues 
 are directly pooled into a single feature vector.
We propose a new zero-shot skeleton-based action recognition method via mutual
information (MI) estimation and maximization.
Specifically, 
1) we maximize the MI between visual and semantic space for distribution alignment; 
2) we leverage the temporal information for estimating the MI 
by encouraging MI to increase as more frames are observed. 
Extensive experiments on three large-scale skeleton action datasets confirm the effectiveness of our method.
\end{abstract}


\maketitle

\section{Introduction}
\label{sec:introduction}

\begin{figure}[!t]
	\begin{center}
		\includegraphics[width=0.9\linewidth]{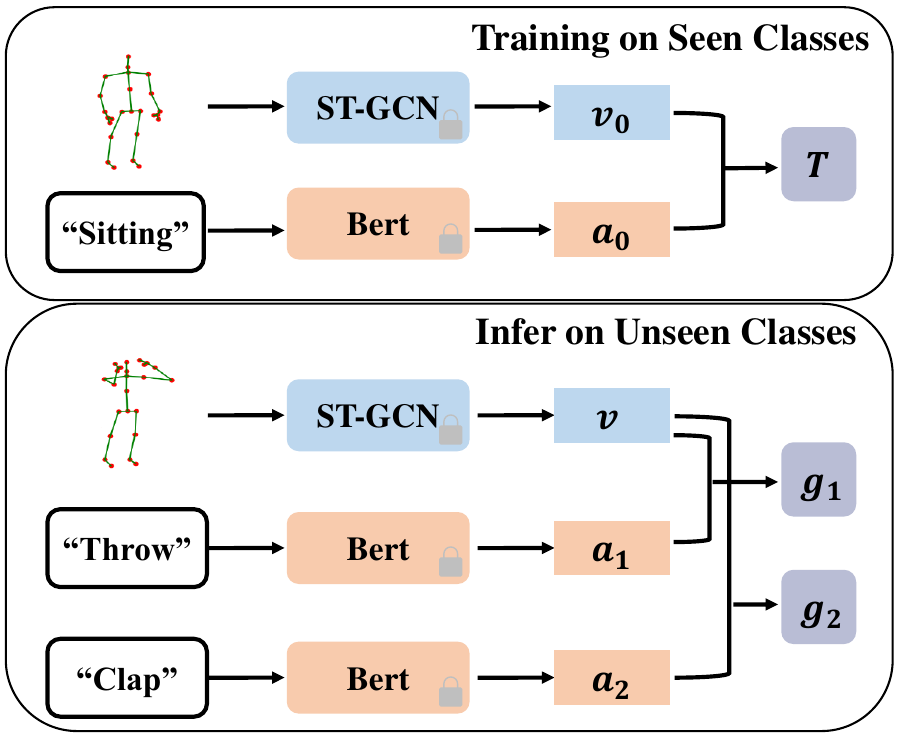}
	\end{center}
	\vspace{-10pt}
	\caption{The core of the zero-shot learning lies in constructing a connection model $T$ between
        visual features $v$ and semantic features $a$ during the training phase. 
    At test time, the learned model $T$ is utilized to predict the most compatible semantic attribute for a given unseen-class visual feature.
	}
	\label{fig:teaser}
	\vspace{-15pt}
\end{figure}

Human action recognition becomes an essential component in many real-world applications, 
including but not limited to security and human-robot interaction. 
It is not hard to acquire skeleton data due to the development of pose estimation~\cite{hua2022weakly} and sensors~\cite{zhang2012microsoft}.
To this end, human skeleton data has emerged as a promising alternative to traditional RGB video data 
due to its robustness to variations in appearance and background, 
as well as its ability to provide an unbiased representation of individuals.

Many researchers explore fully supervised methods for this task, which requires large amounts of labeled training samples. 
However, it is not economical to handle numerous action classes in real-world scenarios since 
the samples of many actions are time-consuming and expensive to collect.
Thus, zero-shot learning~\cite{li2019rethinking,schonfeld2019generalized}
is used to recognize new classes if there are no training samples 
but only some semantic information such as the names, attributes, or descriptions of new classes is available. 
Annotating and labeling 3D skeleton action data presents increased difficulty due to the inclusion of depth information 
and the complexity of human action semantics.
Hence, zero-shot skeleton-based action recognition is highly desirable in 
practical applications because it can significantly reduce the need for collecting and annotating new actions.

In Fig.~\ref{fig:teaser}, Given pre-extracted visual and semantic features,
the core of zero-shot learning is to establish a connection model between visual and semantic spaces in the seen classes.
During the test phase, the learned model is used to facilitate the knowledge transfer from the seen to the unseen classes. 
To address the transfer learning problem, zero-shot action recognition relies on the external knowledge base, i.e., 
the semantic embeddings of each class label from pre-trained large-scale language model such as
Sentence-Bert~\cite{reimers2019sentence-bert} or CLIP~\cite{radford2021learning}.
The effective utilization of semantic information is important for bridging the gap between two different modalities.

There are a few studies on zero-shot skeleton-based action recognition. 
Existing methods~\cite{jasani2019skeleton,gupta2021syntactically} embed action sequences into visual features. 
To establish the connection model of visual and semantic space, a
compatible projection function~\cite{jasani2019skeleton} or a deep metric~\cite{frome2013devise} 
is learned based on the data of seen classes in the training phase. 
Then in the testing phase, the similarities between the visual feature of a test action sequence 
and the sentence embeddings~\cite{jasani2019skeleton} or part-of-speech tagged words~\cite{wray2019fine-grained} of 
the unseen classes are measured either in the projected common space or by the learned metrics. 
However, the projection operation merely maps the visual or semantic features 
to a common anchor point in the embedding space, 
overlooking the global alignment between the distributions of visual and semantic features.
Furthermore, the learned projections or metrics inadequately utilize semantic information 
to capture the associations between the two modalities.
Their attempt to perform cross-modal reconstruction without aligning the distributions 
is challenging due to the significant difference between the visual and semantic spaces, 
ultimately resulting in the difficulty of generalizing to novel classes with diverse distributions.

Secondly, The information loss becomes severe in the zero-shot action recognition task scenario 
since some semantic classes require dynamics information to differentiate from each other.
For example, "walking" and "skipping" differ only in local parts 
since the initial frames for these two action sequences are similar; 
"Skipping" cannot be identified until a human-rising procedure is observed.
Thus, for a human action, utilizing the inherent temporal dynamics information also plays a role in 
the generalization ability of the zero-shot connection model. 

In this paper, we propose a \textbf{S}keleton-based \textbf{M}utual \textbf{I}nformation 
\textbf{E}stimation and maximization framework for 
zero-shot action recognition (SMIE).
To better capture the dependencies between visual and semantic spaces, 
our approach avoids direct mapping and instead aligns the distributions of these two spaces 
using a global alignment module. 
This module utilizes mutual information as a measure of similarity 
and applies an estimator based on Jensen-Shannon divergence (JSD) to 
maximize the mutual information between paired visual and semantic features 
while minimizing mutual information between unpaired visual and semantic features. 
A neural network is employed as the connecting model to 
estimate the similarity score in the JSD estimator, which is used during the test phase on unseen classes.
Then considering the inherent temporal information of actions, SMIE proposes
a temporal constraint module to encourages 
the mutual information between visual and semantic features to increase 
when more parts of the action are executed. 
Specifically, the JSD estimator applies contrastive learning to estimate the global mutual information.
The paired visual and semantic features form positive samples, while unpaired ones form negative samples.
To perceive keyframes that contain more discriminative information in the action sequence,
the temporal constraint module computes the motion attention of each sequence and
masks the keyframes with higher attention to generate extra positive samples, 
which contain partial temporal information loss.
During training, the temporal-constrained mutual information is computed with the same negative samples
and is kept smaller than the global mutual information.

The major contributions of this paper are three-fold:
\vspace{-2pt}
\begin{itemize}
\item We propose a skeleton-based mutual information 
estimation and maximization framework (SMIE), 
a new zero-shot approach to skeleton-based action recognition based on mutual information maximization,
which can capture the complex statistical correlations between the distributions of the visual space and the text semantic space.
\item A novel temporal constraint module is proposed to compute the temporal-constrained mutual information
and a temporal rank loss is applied to help the connection model capture the inherent temporal information of actions.
\item Extensive experiments and analyses demonstrate the effectiveness of the proposed method, 
which outperforms the baseline methods by a large margin. 
\end{itemize}
\vspace{-5pt}
\vspace{-5pt}
\section{Related Work}
\label{sec:related}

{\noindent\bf Zero-shot Action Recognition.}
Most of the existing zero-shot video classification methods aim to build 
the connection between the visual and semantic spaces using feature projections,
which mainly focuses on 
the visual space~\cite{han2020learning,wang2017zero-shot,7410639},
the semantic space~\cite{Brattoli_2020_CVPR,Zhu_2018_CVPR}, 
and the intermediate space~\cite{gan2015exploring,xu2016multi-task}.
Specifically, The visual features are first extracted from videos using a pre-trained network 
such as Convolutional 3D Network (C3D)~\cite{tran2015learning}, ResNet~\cite{he2016deep},
and Inflated 3D Network(I3D)~\cite{carreira2017quo}. 
And then they map the visual or semantic features to the fixed anchor points in the embedding space.
Different from these works, 
we focus on zero-shot skeleton-based action recognition, 
where the visual features for skeleton-based action sequences greatly differ from those for RGB videos. 
We use mutual information instead of projections to associate the skeleton-based visual and semantic features.

\vspace{4pt}
{\noindent\bf Skeleton-based Action Recognition.}
With the development of highly accurate depth sensors such
as Kinect cameras and pose estimation algorithms~\cite{shotton2011real,cao2017realtime}, 
skeleton-based action recognition has attracted increasing attention recently.
Human skeleton-based representation~\cite{guo2022contrastive,Li_2021_CVPR}
is robust to variations of appearance and background environment, 
where each skeleton contains different types of joints,
and each joint records its 3D position. 

Specifically, skeletons are organized as pseudo-images~\cite{ke2017a,kim2017interpretable} 
or a sequence of long-term contextual information~\cite{du2015hierarchical} and feed it into CNNs/RNNs.
Later, ST-GCN~\cite{yan2018spatial} constructs a 
predefined spatial graph based on the natural connections of joints in the human body
and utilizes GCN to integrate the skeleton joint information. 
In terms of consecutive frames, ST-GCN constructs the temporal edges between corresponding joints. 
After that, many variants of ST-GCN are proposed~\cite{shi2019two,zhang2020semantics,cheng2020skeleton-based, wang2023spatio}, 
which contain more data streams or add attention mechanisms. 
In this paper, we employ ST-GCN~\cite{yan2018spatial} and Shift-GCN~\cite{cheng2020skeleton-based} 
as the backbone to extract visual features. 
To explore the temporal relationship of skeleton data, we retain the pre-trained GCN model
and acquire the partial visual feature by inputting frames with different indices in the skeleton sequences. 

\vspace{4pt}
{\noindent\bf Zero-shot Skeleton-Based Action Recognition.}
Fewer works have been devoted to zero-shot skeleton-based action recognition though it is of great importance. 
DeViSE~\cite{frome2013devise} and RelationNet~\cite{jasani2019skeleton} are extended to tackle this problem 
by extracting visual features from skeleton sequences with ST-GCN 
and semantic embeddings with Word2Vec~\cite{mikolov2013distributed} or
Sentence-Bert~\cite{reimers2019sentence-bert}. 
DeViSE uses a simple learnable linear projection between the visual and semantic feature spaces. 
Based on it, RelationNet utilizes an attribute network and a relation network to achieve the same goal. 
In the above works, a projection operation is needed to build the visual-semantic embedding, 
they map the visual or semantic features to the fixed anchor points in the embedding space, 
which does not consider the global distribution of the semantic features.
Recently, SynSE~\cite{gupta2021syntactically} uses a generative multi-modal 
alignment module to align the visual features with parts of speech-tagged words. 
It works but needs extra PoS syntactic information to divide labels into verbs and nouns. 
Our method differs from these works in two aspects: 
We maximize the mutual information between the two modalities, 
and the distribution between visual and semantic features can be aligned. 
The global information of the semantic distribution can be utilized, 
which can help to guide the knowledge transfer from seen domain to the unseen domain. 
Second, to exploit the temporal information of the skeleton sequence, 
a temporal constraint module is used to encourage the mutual information between visual 
and semantic features to increase with the number of observed frames.

\vspace{4pt}
{\noindent\bf Mutual Information for Zero-shot Learning.}
Recently, the mutual information between the visual space and the semantic space 
has been explored in zero-shot learning. 
In~\cite{sylvain2020locality}, 
local patches from other images with the same label 
are drawn to estimate the mutual information for local interpretability. 
In~\cite{tang2020zero-shot}, 
mutual information is utilized to learn latent visual and semantic representations 
so that multi-modalities can be aligned for generalized zero-shot learning. 
Different from the above,
we use mutual information to bridge the skeleton-based visual space and the text label semantic space 
and impose a novel constraint on the mutual information 
to capture the temporal semantic of the skeleton sequences. 
\section{Method}
\label{sec:method}

\begin{figure*}[!ht]
	\begin{center}
	\includegraphics[width=0.9\textwidth]{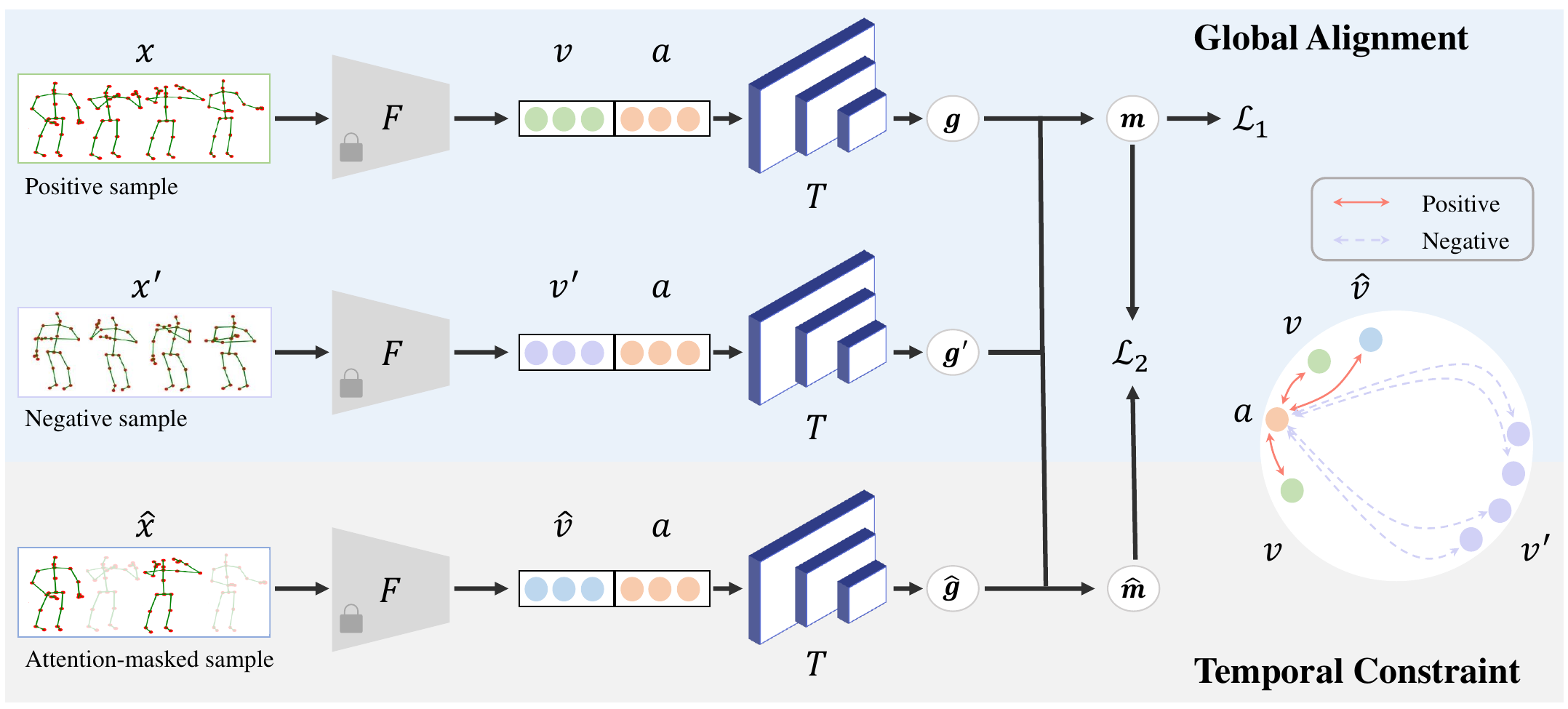}
        \vspace{-10pt}
        \caption{The framework of our proposed SMIE includes a global alignment module to estimate the mutual information between visual and semantic features, and a temporal constraint module to incorporate temporal dynamics information into the estimation network.}
	\label{fig:mi}
	\vspace{-15pt}
    \end{center}
\end{figure*}

\subsection{Problem Definition}
The zero-shot learning setting addressed in this paper is the same as~\cite{jasani2019skeleton}, where
the model is trained on seen classes and tested on disjoint unseen classes.
Specifically, the training dataset consists of the skeleton sequence and the corresponding class name from \textbf{seen} classes.
Each training sample is denoted as $(x^s, e^s)$, ${x^s} \in \mathbb{R}^{K \times J \times C}$ represents the training skeleton sequence with
$K$ frames and
$J$ recorded 3D-joints for each frame, and ${e}^s$ is the corresponding class name.
The test dataset comes from \textbf{unseen} classes, a sample of the test dataset is denoted as
$(x^u, e^u)$. 

Generally, we use a visual feature extractor $F_v$, which has been pre-trained on seen classes, 
and a semantic feature extractor
$F_e$, which has been pre-trained on large-scale language models.
These two extractors are employed to acquire the visual and semantic features $v$ and $a$ 
by taking in the skeleton sequence and class name as inputs.
Then a layer-norm layer (without learnable parameters) $N$ is used to normalize the visual features,
\vspace{0pt}
\begin{equation}
v^i = N(F_v(x^i)), \, a^i = F_e(e^i), \,  i \in \left\{ {s,u} \right\}.    
\end{equation}
\vspace{0pt}
Let $V$ and $A$ represent the random variables of $v$ and $a$ respectively. 
The zero-shot learning aims 
to classify the sample of the unseen classes by the model 
learned based on the training data from seen classes. 
The utilization of semantic features is important because the semantic space is shared between seen and unseen classes, 
which can help the learned model transfer knowledge from different domains. 
For simplicity, we omit the subscript for seen (s) and unseen (u) classes 
during the following model training introduction.

\subsection{Method Overview}
In Fig.~\ref{fig:mi},
we propose a skeleton-based mutual information 
estimation and maximization framework (SMIE).
Our SMIE consists of two modules: 
The global alignment module uses mutual information estimation and maximization
to capture the statistical correlations between visual and semantic distributions.
The temporal constraint module is utilized to enable the connection model $T$ to perceive the keyframes of sequences,
which allows for the exploration of action dynamics and the capture of inherent temporal information.

In the inference phase, 
the trained mutual information estimation network $T$ calculates the similarity score $g$
between tested visual sequences and all semantic features of unseen classes. 
The unseen class with the highest similarity score is chosen as the prediction.

\subsection{Global Alignment}
\label{sec:global}
Previous zero-shot skeleton-based action recognition approaches~\cite{frome2013devise,jasani2019skeleton} 
learn projections that pull the visual features and the corresponding semantic features closer in seen classes 
without considering the whole distribution of features. 
Due to the huge gap between  visual and semantic spaces, 
it is difficult to bridge the cross-domain gap and generalize the projections to unseen classes. 

To tackle this issue, we design a global alignment module to 
learn an estimation network $T$ through maximizing the mutual information $I(V; A)$~\cite{tschannen2020on, qiang2021robust} 
between the random variables of visual and semantic features:
\begin{equation}
\begin{split}
I(V; A) & = D_{KL}( p(v, a) || p(x) p(a) ) \\
& = \mathbb{E}_{p(v, a)}
\begin{bmatrix}
\log \frac{p(v \mid a)}{p(v)}
\end{bmatrix}.
\end{split}
\label{eq:MI}
\end{equation}
\vspace{1pt}
$D_{KL}$ is the KL-divergence between $p(v, a)$ and $p(v) p(a)$, 
which represents the joint distribution and product of the marginal distributions of $v$ and $a$, respectively. 
The information on joint distribution can be utilized, 
and help the model use the global semantic attributes. 
Meanwhile, the mutual information can be written with the joint entropy among~$V$, $A$ and their conditional entropy,
\vspace{2pt}
\begin{equation}
I(V ; A)=H(V, A)-(H(V \mid A)
+H(A \mid V)).
\label{eq:entropy}
\end{equation}
\vspace{2pt}
As pointed out above, maximizing the mutual information between $V$ and $A$
is equivalent to maximizing the common information, i.e.,
the difference between joint entropy and conditional entropy.

Following Eq.~\eqref{eq:MI}, 
instead of directly modeling $p(v \mid a)$, 
mutual information is utilized to encode $V$ and $A$ 
into compact distributed vector representations via a connection network learned from data. 
Benefiting from it, the captured shared information 
between visual and semantic features maintains a better global structure,
while low-level information and noise will be discarded. 

However, directly calculating the mutual information of 
two random variables in high-dimensional spaces is extremely hard.
Inspired by the Jensen-Shannon divergence (JSD)~\cite{nowozin2016f-gan}, 
we propose to learn an estimation network $T$ 
that takes the global visual feature $v$ and the semantic feature $a$ 
as inputs. 
The output of the network can be served as a similarity metric score $g$
between the visual and semantic features, 
which is applied to match a skeleton sequence to 
the unseen classes in the testing phase.

The estimation network $T$ can be trained by maximizing the following JSD estimator. 
\begin{equation}
\begin{split}
I(V;A) \approx m = \mathbb{E}_{p(v, a)}[-f_{sp}(-T(v,a))] \\
-\mathbb{E}_{p(v)p(a)}[f_{sp}(T(v^{\prime},a))],
\end{split}
\label{eq:jsd}
\end{equation}
\vspace{2pt}
where $m$ is the estimated mutual information shown in Fig.~\ref{fig:mi}.
Note that $(v, a)$ are paired visual/semantic features 
and $(v^{\prime}, a)$ are negative pairs.
Specifically, $v$ is a visual feature extracted from skeleton sequences $x$, 
which is related to $a$.
And $v^{\prime}$ is a visual feature extracted from negative sample $x^{\prime}$ related to other classes. 
$f_{sp}$ is the soft-plus function $f_{sp}(z) = \log(1+e^{z})$. 
Then, the visual feature $v$ is concatenated 
with the paired semantic feature $a$ to form the positive pair, 
while the negative pair is generated by concatenating this semantic feature 
with the visual feature $v^{\prime}$ of another sequence. 
The two pairs are sent to the estimation network $T$ to obtain the scores $g$ and $g^{\prime}$, 
respectively, for contrastive learning.

In this way, the estimation network encourages the semantic feature 
to have a larger similarity to the corresponding visual feature than those unpaired visual features. 
Thus, to maximize the estimated mutual information $m$ for the global alignment, 
we have the following loss:
\vspace{2pt}
\begin{equation}
\mathcal{L}_1 = -m.
\end{equation}

The parameters of the estimation network $T$ are updated by gradient descent during training.

\subsection{Temporal Constraint Module}
\label{sec:temporal}
Compared with the image data, 3D human skeleton data is more complicated 
because of the additional temporal dimension. 
Utilizing the temporal dynamics information within a skeleton sequence 
can help the model capture subtle differences among various classes.
A temporal constraint module is proposed to incorporate such temporal information.

Generally, for human action sequences, 
the more frames observed the more dynamics information model can capture, 
which encourages the visual feature to get a stronger correlation 
with its corresponding semantic feature. 
Furthermore, the keyframes in the action sequences are often richer in discriminative information,
so the sequence loss of such frames has lower semantic relevance to their labels.
Inspired by PSTL~\cite{zhou2023self}, which utilizes the motion of skeleton data to find the key frames of each sequence.
We further adopt bidirectional motion attention to enhance the effectiveness.
As shown in the bottom half of Fig.~\ref{fig:mi}, 
to acquire the attention-masked sample, we first calculate the bidirectional action attention for each action sequence.
Specifically, the motion $p \in \mathbb{R}^{K \times J \times C}$ of the sequence is computed by the temporal displacement between frames:
$p_{k,j,c}^{nex} = x_{k+1,j,c}-x_{k,j,c}$, which represents the subsequent variation of action in each frame.
Then we further incorporate the displacement between the current frame and its preceding frame into the motion information:
$p_{k,j,c}^{pre} = x_{k-1,j,c}-x_{k,j,c}$.
The bidirectional motion of the sequence can be defined as:
\vspace{2pt}
\begin{equation}
p_{k,j,c} = (p_{k,j,c}^{nex})^2 + (p_{k,j,c}^{pre})^2.
\label{eq:motion}
\end{equation}
\vspace{2pt}
Then, we calculate the average motion value for each frame to acquire $p_k$:
\begin{equation}
p_k=\frac{1}{J \times C} \sum_{j=1}^J \sum_{c=1}^C p_{k, j, c}.
\label{eq:motion}
\end{equation}
With the bidirectional motion $p_{k}$, we can acquire the overall motion rate of a frame which serves as 
the bidirectional attention weight:
\vspace{2pt}
\begin{equation}
q_k = \frac{p_k} {\sum_{i=1}^{K} {p_i}}.
\end{equation}
Then the top $P$ frames with the highest attention scores $q_{k_1},...,q_{k_P}$ are selected
and the frame list $x_{k_1},...,x_{k_P}$ serves as the keyframes which contain more discriminative information about the action.
We mask such key frames on the original skeleton sequence $x$ to construct the attention-masked sample
sequence $\hat{x}$, which suffers some information loss and has lower semantic relevance to their semantic feature.
With the help of the visual feature extractor $F$ and layer-norm layer $N$, 
the temporal-constrained visual feature $\hat{v}$ is extracted and concatenated with the corresponding semantic 
feature $a$ to construct the  temporal-constrained positive pair. 

Similar to the usage of the JSD estimator in the global alignment module, 
the temporal-constrained mutual information $\hat{m}$ is formulated as follows, 
\vspace{2pt}
\begin{equation}
\begin{split}
\hat{m} = \mathbb{E}_{p(\hat{v}, a)}[-f_{sp}(-T(\hat{v},a))] \\
-\mathbb{E}_{p(\hat{v})p(a)}[f_{sp}(T(v^{\prime},a))],
\end{split}
\label{eq:jsd2}
\end{equation}
\vspace{2pt}
Here, the mutual information 
between the temporal-constrained visual features and corresponding semantic features is maximized. 
Note that the negative pair still consist of the original negative sample $x^{\prime}$ 
and the unpaired semantic feature $a$ to ensure the consistency of the negative sample space.
Our temporal constraint module aims to encourage the connection module to perceive the importance of the keyframes during  
mutual information estimation.
So a hinge loss is utilized to force the global mutual information $m$ 
greater than the partial one during training, 
\vspace{2pt}
\begin{equation}
\begin{split}
\mathcal{L}_2 = \mathrm{max}(0, \beta -(m-\hat{m})),
\end{split}
\label{eq:hinge}
\end{equation}
\vspace{2pt}
where $\beta$ is a hyper-parameter to control the distance between two types of 
mutual information. 
By adjusting $\beta$, the model can adapt to different datasets.
In brief, the temporal constraint module serves as a regularization on the JSD estimator,
which helps the model incorporate the dynamics information
and be more robust.

The overall loss function combines the global mutual information maximization term 
and the temporal constraint term together, as shown in the following, 
\begin{equation}
\mathcal{L} =  \mathcal{L}_1 + \lambda \mathcal{L}_2.
\label{eq:loss}
\end{equation}
$\lambda$ is the trade-off parameter and is set to $0.5$ for all experiments.

\section{Experiments}
\label{sec:experiment}

\subsection{Datasets}

\vspace{4pt}
\noindent\textbf{NTU-RGB+D 60}~\cite{shahroudy2016ntu} contains $56,578$ skeleton sequences of $60$ action categories,
performed by $40$ volunteers.
The skeleton sequences are collected by Microsoft Kinect sensors and each subject is represented by $25$ joints. 
Two official dataset splits are applied: 
1) Cross-Subject (xsub): the training set contains half of the subjects, 
and the rest make up the testing sets;
2) Cross-View (xview): The data from different views constitute the training and test set. 

\vspace{4pt}
\noindent\textbf{NTU-RGB+D 120}~\cite{liu2019ntu} is the extended version of the NTU-60.
It is performed by $106$ volunteers and
contains $113,945$ skeleton sequences of $120$ action categories.
NTU-120 also has two official dataset splits:
1) Cross-Subject (xsub): $53$ subjects belong to the training set
and the testing data is performed by the rest volunteers;
2) Cross-Setup (xset): the training set is captured by cameras with even IDs
and the test set is captured with odd IDs.

\vspace{4pt}
\noindent\textbf{PKU-MMD}~\cite{liu2020benchmark} has almost $20000$ action samples 
in $51$ categories collected by $66$ subjects.
It is captured via the Kinect v2 sensors from multiple viewpoints.
The dataset has two parts:
1) Part I contains $21539$ samples;
2) Part II contains $6904$ samples.

\begin{table}[!t]
\begin{center}
\caption{The hyper-parameters on NTU-60, NTU-120, and PKU-MMD datasets.}
\vspace{-10pt}
\label{tab:hypara}
\begin{tabular}{l|ccc}
\toprule
Parameter & NTU-60 & NTU-120 & PKU-MMD \\ \midrule
$\beta$         & 0.1    & 0.5    & 0.01     \\
$P$         & 15      & 15     & 15      \\ \bottomrule
\end{tabular}
\end{center}
\vspace{-15pt}
\end{table}

\subsection{Implementation Details and Baselines}

\vspace{4pt}
\noindent\textbf{Detailed Implementation of SMIE.}
We follow the same data processing procedure in Cross-CLR~\cite{Li_2021_CVPR}, 
which removes the invalid frames and resizes the skeleton sequences to 50 frames by linear interpolation.
ST-GCN~\cite{yan2018spatial} with 16 hidden channels is used as the visual feature extractor 
and the extracted feature dimension is $256$. 
For the semantic feature, we use Sentence-Bert~\cite{reimers2019sentence-bert} 
to obtain the 768-dimensional word embeddings, and then
all semantic features are processed by L2 normalization to improve the stability of the training phase.
For all experiments, we adopt the Adam optimizer and the CosineAnnealing scheduler with 100 epochs.
The mini-batch size is 128.
The learning rate is $1e-5$ for NTU-60 and PKU-MMD datasets, 
while for the NTU-120 dataset with larger data size, the learning rate is $1e-4$.
Tab.~\ref{tab:hypara} shows the choices of the hyper-parameters for all datasets.
$P$ refers to the number of masked keyframes, which remains $15$ for all three datasets.
The hyper-parameter margin $\beta$ controls the distance between global and temporal-constrained mutual information.
By adjusting $\beta$, the model can adapt to different datasets. 
Note that, with decreasing margins, the impact of temporal constraints on the overall loss increases.
Table~\ref{tab:hypara} indicates a positive correlation between dataset size
and margin parameter $\beta$. Smaller datasets necessitate a smaller $\beta$ for optimal performance.
The reason is that the temporal constraint module serves as a regularization on the JSD estimator.
By providing necessary constraints on the model, the module prevents overfitting to the limited dataset.

For the details of our SMIE model, the network $T$ in Eq.~\eqref{eq:jsd} 
is composed of 
three MLP layers with ReLU activation functions. 
For the negative pairs, we shift the skeleton visual samples in a batch to make the visual and 
semantic features do not correspond.

\vspace{4pt}
\noindent\textbf{Baseline Methods.}
The core of skeleton-based zero-shot learning lies in evaluating the efficacy of the connection model,
which functions as the intermediary between the visual and semantic spaces.
The recent method SynSE~\cite{gupta2021syntactically} follows this main idea and provides various zero-shot learning comparative methods, such as Devise~\cite{frome2013devise}, ReViSE~\cite{tsai2017learning}, RelationNet~\cite{jasani2019skeleton}, JPoSE~\cite{wray2019fine-grained} and CADA-VAE~\cite{schonfeld2019generalized}.
Specifically, DeViSE and RelationNet both use linear projections to map visual features and
semantic features into the same space. 
After the projection, 
DeViSE calculates the dot-product similarities between projected visual and semantic features.
RelationNet utilizes a relation module to acquire the similarity between projected features. 
ReViSE uses a maximum mean discrepancy loss as cross-domain learning criteria to align the latent embeddings. 
JPoSE performs cross-modal fine-grained action retrieval between text and skeleton data. 
It learns PoS-aware embeddings and builds a separate multi-modal space for each PoS tag. 
CADA-VAE learns a latent space for both visual features and semantic embeddings 
via aligned variational autoencoders. 
Based on the above, the state-of-art method SynSE
infuses latent skeleton visual representations with PoS syntactic information.
We conducted an apple-to-apple comparison between our SMIE and such baseline methods.

\subsection{Comparison with State-of-the-Art}

\begin{table}[t]
\caption{Comparison of SMIE with the State-of-the-Art methods on NTU-60 and NTU-120 datasets.}
\vspace{-5pt}
\label{tab:sota}
\begin{center}
\begin{tabular}{l|cc|cc}
\toprule
Method & \multicolumn{2}{c|}{NTU-60(\%)} & \multicolumn{2}{c}{NTU-120(\%)} \\
Split    & 55/5  & 48/12 & 110/10 & 96/24 \\\midrule

DeViSE          & 60.72        & 24.51      & 47.49        & 25.74       \\
RelationNet     & 40.12       & 30.06       & 52.59        & 29.06       \\
ReViSE          & 53.91       & 17.49       & 55.04        & 32.38       \\
JPoSE          & 64.82       & 28.75       & 51.93        & 32.44       \\
CADA-VAE        & 76.84       & 28.96       & 59.53        & 35.77       \\
SynSE           & 75.81       & 33.30       & 62.69        & 38.70       \\\midrule
\textbf{SMIE}                   & \textbf{77.98}       & \textbf{40.18}       & \textbf{65.74}        & \textbf{45.30}       \\\bottomrule
\end{tabular}
\end{center}
\vspace{-8pt}
\end{table}

\vspace{4pt}
\noindent\textbf{Evaluation Settings.}
In zero-shot learning, the selection of distinct class splits gives rise to varied sets of seen and unseen classes, 
exerting a significant impact on the empirical outcomes.
Moreover, the accuracy is also affected by the selection of the feature extractor.
To facilitate a direct comparison with the state-of-the-art approach and fully demonstrate the 
effectiveness of our connection model, we employ identical experimental settings as SynSE.
That means we use the same class splits, and pre-extracted visual and semantic features as supplied in their codebase.
Specifically, SynSE maintains two types of fixed class splits on NTU-60 and NTU-120 datasets.  
For the NTU-60 dataset, SynSE provides 55/5 and 48/12 splits, which include 5 and 12 unseen classes, respectively. 
Meanwhile, for the larger NTU-120 dataset with more categories, SynSE offers 110/10 and 96/24 splits.
The visual feature extractor employed in the study is Shift-GCN~\cite{cheng2020skeleton-based}, 
while the semantic feature extractor utilized is Sentence-Bert~\cite{reimers2019sentence-bert}.

\vspace{4pt}
\noindent\textbf{Results and Analysis.}
Tab.~\ref{tab:sota} presents the comparison results with baseline methods on NTU-60 and NTU-120. 
For the 55/5 and 48/12 split on NTU-60 datasets, 
SMIE outperforms SynSE $2.17$\% and $6.88$\% (relatively $2.86$\% and $20.66$\%), respectively. 
For the 110/10 split and 96/24 split on NTU-120, 
SMIE achieves $3.05$\% and $6.60$\% (relatively $4.87$\% and $17.05$\%), respectively. 
As the number of unseen classes increases, 
the difficulty of generalizing the learned knowledge from seen to unseen classes also increases.
Notably, Our proposed SMIE method utilizes mutual information to capture global semantic information 
and effectively bridges the gap between visual and semantic space, 
which shows promising potential in enhancing the performance of unseen classes.

\subsection{Ablation Study}

\vspace{4pt}
\noindent\textbf{Optimized Experimental Setting.}
From the experimental setting of SynSE, 
we find the objective of zero-shot learning experiments is to verify the effectiveness of the learned connection model.
However, due to the significant impact of different class splits on the results, 
there can be a considerable deviation in accuracy even if the number of unseen classes is the same.
Meanwhile, it is advisable to minimize the impact of the feature extractors with complex structures on the results, 
in order to focus on the effectiveness of the connection model itself.
Thus, we provide an optimized experimental setting for zero-shot skeleton-based action recognition.
First, we expand the dataset from two to three large-scale skeleton datasets, i.e., NTU-60, NTU-120, and PKU-MMD datasets,
which increases the credibility of the results.
Second, for each dataset, a three-fold test is applied to eliminate variance. 
Each fold has different groups of seen and unseen classes and the average results are reported.
At last, we follow most skeleton-based self-supervised methods~\cite{zhou2023self,Li_2021_CVPR,lin2020ms2l,thoker2021skeleton} 
and apply the classical ST-GCN~\cite{yan2018spatial} as the visual feature extractor to minimize the impact of the feature extractors.
The semantic feature extractor utilized in this study is Sentence-Bert, which is consistent with SynSE.


\begin{table}[!t]
\begin{center}
\caption{Ablation studies under optimized experimental setting on NTU-60, NTU-120, and PKU-MMD datasets.}
\vspace{-5pt}
\label{tab:overall}
\resizebox{0.95\linewidth}{!}{%
\begin{tabular}{l|c|c|c}
\toprule
Method                           & NTU-60(\%)       & NTU-120(\%)    &PKU-MMD(\%)\\
Split                           & 55/5       & 110/10    & 46/5\\
\midrule
DeViSE                   & 49.80        & 44.59     & 47.94\\
RelationNet              & 48.16        & 40.55     & 51.97\\
ReViSE                   & 56.97        & 49.32     & 65.65\\\midrule
SMIE w/o $\mathcal{L}_2$  & 62.17        & 55.34     & 66.14  \\
\textbf{SMIE }        & \textbf{63.57}  & \textbf{56.37} & \textbf{67.15}  \\\bottomrule
\end{tabular}
}
\end{center}
\vspace{-15pt}
\end{table}

\vspace{4pt}
\noindent\textbf{Overall Analysis on Optimized Experimental Setting.}
Under the optimized experimental setting, we aim to conduct an ablation study on the temporal constraint module of the SMIE.
To provide more baseline results for subsequent research work, 
we reproduced two mapping methods (DeViSE and RelationNet) and one distribution alignment method (ReViSE) under this setting.
Specifically, for the NTU-60 and PKU-MMD datasets,
5 unseen classes are selected randomly and the rest serves as the seen classes.
The visual extractor only pre-trains on the seen classes.
For the NTU-120 datasets, the number of unseen classes is 10. 
All the datasets get 3 groups of random splits and 
the average results are reported on Tab.~\ref{tab:overall}.
We found out that the projection-based methods
obtain relatively lower accuracy. 
By aligning the distributions of the latent embeddings in the two domains, 
ReViSE achieves some improvements on the three datasets,
which further confirms the importance of global information to the semantic features. 
However, ReViSE is much more complicated for the utilization of extra auto-decoders.
Our SMIE achieves significant improvements on all datasets. 
Specifically, SMIE outperforms other projection methods by a margin of about 13.77\%, 11.78\%, 
and 15.18\% on the three datasets. 
For ReViSE, our SMIE still achieves 6.60\%, 7.05\%, and 1.50\% increments. 
By utilizing mutual information as the similarity metric, SMIE aligns the distributions of the 
two modalities and incorporates more discriminative information between different features in Eq.~\eqref{eq:entropy}.
For the ablation study of the temporal constraint module,
``SMIE (w/o $\mathcal{L}_2$ )'' in the table indicates that the model with the global alignment module only,
and ``SMIE'' is the full model. 
It is observed the temporal constraint module brings about
$1.40\%$, $1.03\%$, and $1.01\%$ performance on the three datasets, respectively. 
The results show the temporal constraint can help to integrate useful temporal information.

\begin{table}[!t]
\caption{Comparisons of different margin $\beta$ in SMIE on NTU-60, NTU-120, and PKU-MMD datasets.}
\label{tab:para}
\vspace{-10pt}
\begin{center}
\begin{tabular}{l|c|c|c}
\toprule
$\beta$     & NTU-60 (\%) & NTU-120 (\%) & PKU-MMD (\%) \\ \midrule
0        &  62.17 &   55.34     &  66.14       \\
0.01     &  62.98  &  55.93 &   \textbf{67.15 }     \\
0.1      &  \textbf{63.57}  &  55.78 &   66.77      \\
0.5      &  63.29  &  \textbf{56.37} &   66.29      \\
1        &  63.12  &  54.98 &   65.99      \\ \bottomrule
\end{tabular}
\end{center}
\vspace{-15pt}
\end{table}

\vspace{4pt}
\noindent\textbf{Influence of Hyper-parameters.}
To determine the best choice of the margin $\beta$ in SMIE,
we differ it based on our full model and conduct a test with 
all three splits on NTU-60, NTU-120, and PKU-MMD datasets. 
Tab.~\ref{tab:para} shows the overall results.
Note that, with decreasing margins, the impact of temporal constraints on the overall loss increases.
It is found that as $\beta$ increases, the performance first increases and then drops. 
The NTU-60, NTU-120, and PKU-MMD datasets achieve the best results at $\beta$ values of 0.1, 0.5, and 0.01, respectively.
The choice of margin keeps a balance between the global alignment module and the temporal constraint module.
When the $\beta=1$, the temporal constraint can not be fully used, which results in a performance drop. 

\begin{table}[!t]
\begin{center}
\caption{Results of CLIP semantic feature extractor on NTU-60, NTU-120, and PKU-MMD datasets.}
\vspace{-10pt}
\label{tab:clip}
\resizebox{0.95\linewidth}{!}{%
\begin{tabular}{l|c|c|c}
\toprule
Method                           & NTU-60(\%)       & NTU-120(\%)    &PKU-MMD(\%)\\
Split                           & 55/5       & 110/10    & 46/5\\
\midrule
DeViSE                   & 56.61        & 41.55     & 61.72\\
RelationNet              & 56.12        & 32.68     & 56.96\\
ReViSE                   & 55.70        & \textbf{46.72}     & 66.61\\\midrule
\textbf{SMIE }        & \textbf{61.11}  & 45.74 & \textbf{71.50}  \\\bottomrule
\end{tabular}
}
\end{center}
\vspace{-12pt}
\end{table}

\vspace{4pt}
\noindent\textbf{Ablation Studies on Different Semantic Features.}
Under the optimized experimental setting, we also
explored the influence of different semantic feature extractors on experimental results.
As shown in Table~\ref{tab:clip}, we use CLIP~\cite{radford2021learning} as 
the semantic feature instead of Sentence-BERT~\cite{reimers2019sentence-bert}.
Similarly, with different semantic features, the global align methods still achieve better 
performances than the direct mapping methods.
Our proposed SMIE can also outperform the
baseline methods by a large margin on NTU-60 and PKU-MMD datasets. 
On the NTU-120 dataset with more data, our SMIE method achieves results 
comparable to the more complex ReViSE, which utilizes extra visual and textual auto-decoders.
These results demonstrate that our SMIE method can achieve good performance 
on different semantic feature extractors, with a simple and efficient structure.

\begin{table}[!t]
\begin{center}
\caption{Results with the expanding category descriptions by ChatGPT on NTU-60, NTU-120, and PKU-MMD datasets.}
\vspace{-10pt}
\label{tab:chat}
\begin{tabular}{l|c|c|c}
\toprule
Method          & NTU-60 (\%) & NTU-120 (\%) & PKU-MMD (\%) \\ \midrule
SMIE            &  63.57 & 56.37     & 67.15 \\
SMIE\_Chat       &  70.21 & 58.85     & 69.26  \\ \bottomrule
\end{tabular}
\end{center}
\vspace{-15pt}
\end{table}

\vspace{4pt}
\noindent\textbf{Expanding Category Descriptions using ChatGPT.}
Conventional approaches typically rely on taking the category label
as input to a semantic feature extractor, to obtain the corresponding semantic feature.
However, these labels contain only a few words and can not
fully and accurately describe the corresponding action semantics.
Based on it, we expand each action label name 
into a complete action description using ChatGPT
and then extract its semantic feature.
For example, "Wear jacket" can be expanded to "the act of putting on a garment designed to cover the upper body and arms".
Following our optimized experimental setting, the results are shown in Tab.~\ref{tab:chat}.
Significant improvement in SMIE\_Chat indicates that a more comprehensive description of action semantics 
leads to the improved representational capacity of semantic features, 
facilitating the connection model to capture the relationship between visual and semantic spaces.

\begin{figure*}[!t]
\vspace{0pt}
	\begin{center}
		\includegraphics[width=0.80\linewidth]{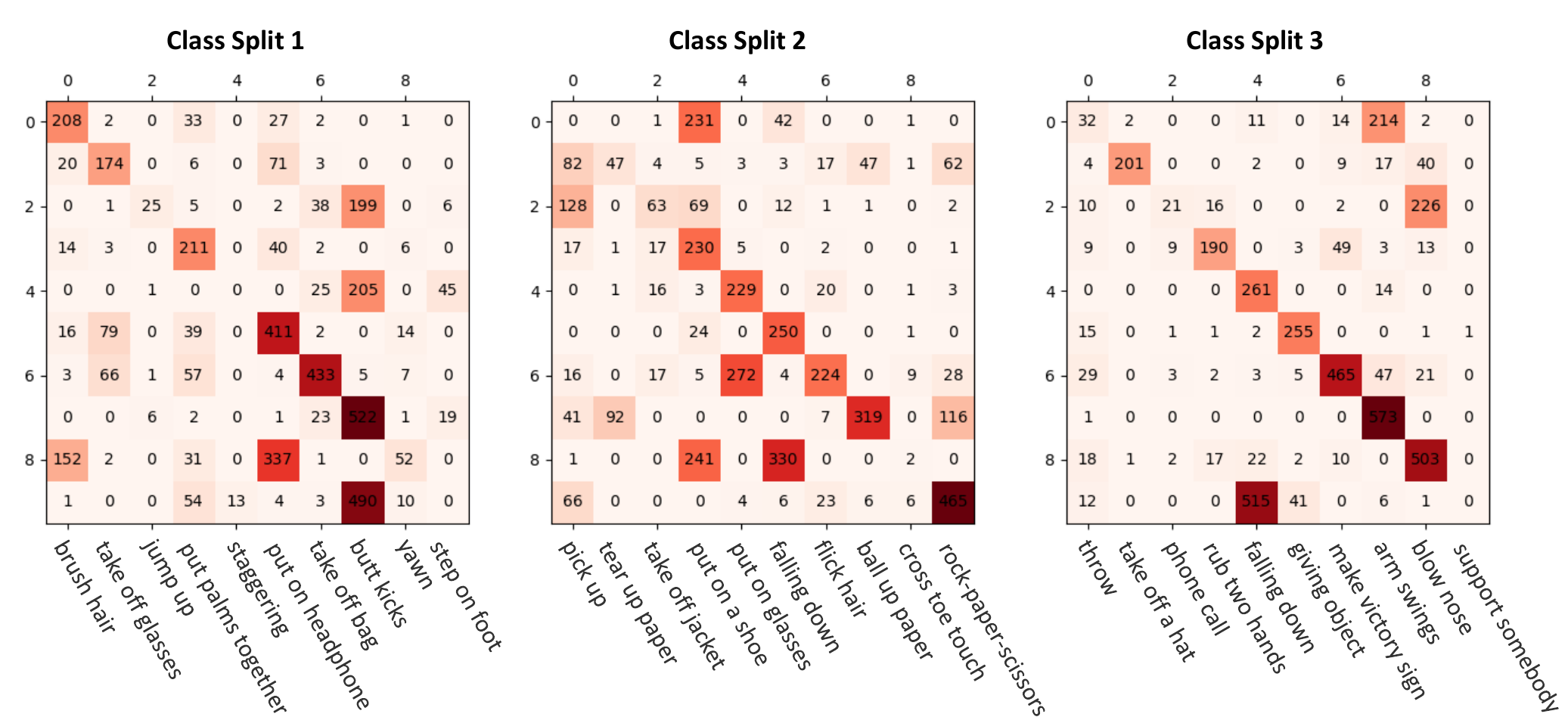}
	\end{center}
	\vspace{-12pt}
	\caption{Confusion matrices for 3 randomly selected class splits on the NTU-120 dataset. The x-axis indicates the predicted class and the y-axis indicates the true class.
	}
	\label{fig:conf}
	\vspace{-12pt}
\end{figure*}

\begin{figure}[!t]
	\begin{center}
	\includegraphics[width=0.9\linewidth]{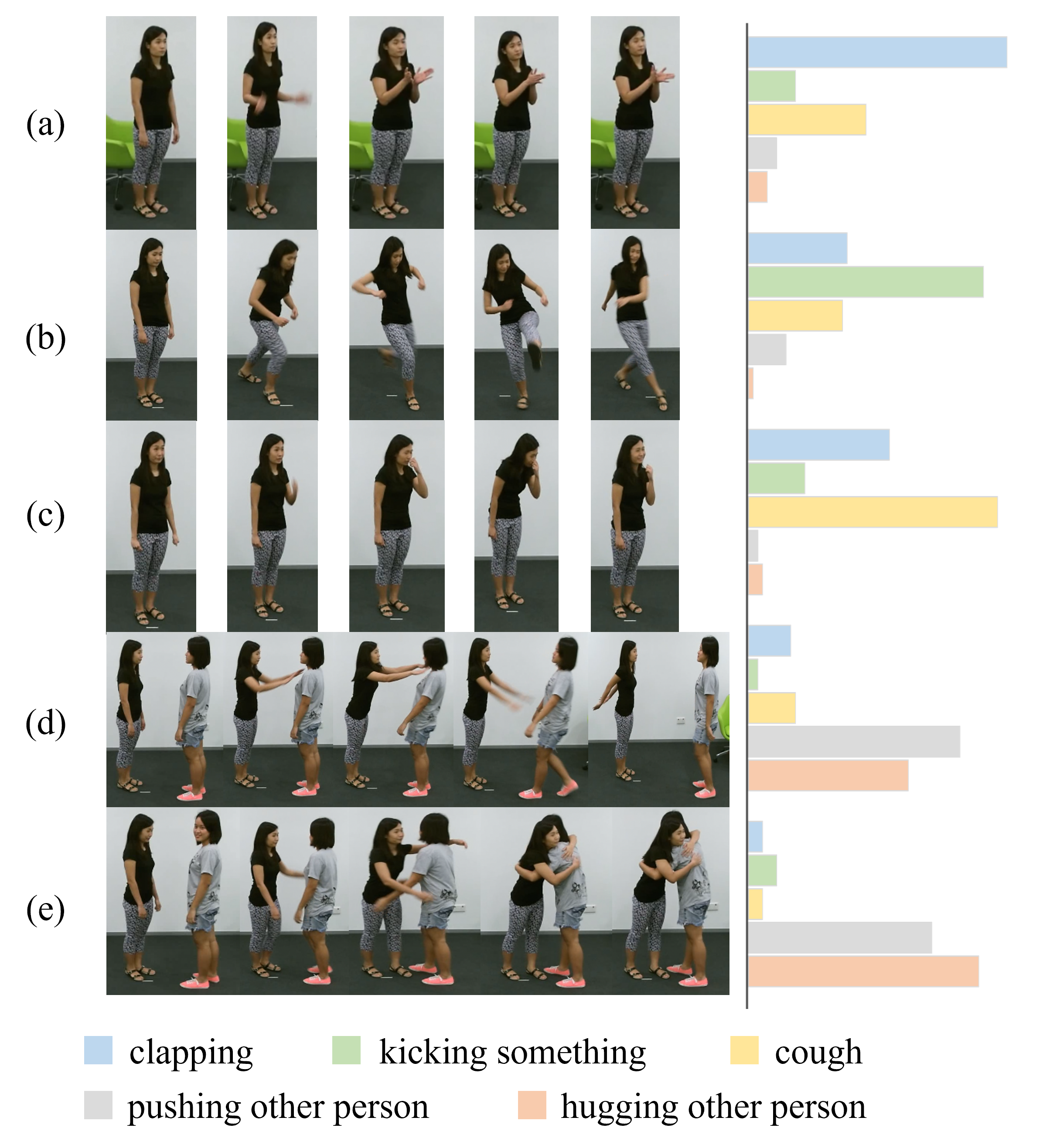}
    \vspace{-12pt}
	\caption{ Visualization of the predicted results on NTU-60 dataset, 
    where the estimated score of each class is represented by the corresponding bar length.
    Best viewed in color.
	}
	\label{fig:qual}
	\vspace{-17pt}
    \end{center}
\end{figure}

\vspace{4pt}
\noindent\textbf{Qualitative Results and Analysis.}
We visualize some predictive scores of the four unseen classes given by our approach in Fig.~\ref{fig:qual}.
As for the one-subject scenario shown in Fig.~\ref{fig:qual} (a, b, c), 
the actions of "clapping" and "cough" are very similar to each other. 
Our model not only makes the right predictions but also 
predicts plausible predictions for similar classes. For example,  
the class with the second highest score is "cough" when the ground truth is "clapping". 
A similar conclusion could be made in the two subjects' scenarios as shown in Fig.~\ref{fig:qual} (d, e),
which proves the rationality of our method.

\vspace{4pt}
\noindent\textbf{Visualization of the Confusion Matrices.}
Fig.~\ref{fig:conf} shows the confusion matrices of 3 random class
splits on the NTU-120 dataset. 
Each matrix contains 10 unseen classes and the number in the matrix represents 
the classified samples. 
By reading the confusion matrices, 
the classification accuracy and misclassification of each class can be understood.
We observe 
1) short actions such as "jump up" or "yawn" are likely to be misclassified 
and the temporal constraint is less effective for them; 
2) some actions such as "pick up" and "tear up paper" are hard to classify 
since they are mostly done with volunteers' hands. 
Thus, we will further investigate these two issues in our future work.

\vspace{4pt}
{\noindent\bf Explanation of Contrastive Learning in SMIE.}
The global alignment module uses contrastive pairs to estimate the mutual information 
between visual and semantic features\cite{hjelm2018learning,qiang2022interventional}. 
It can be motivated by aligning the two distributions of $p(v)$ and $p(a)$ 
to better connect the visual and semantic spaces.
To this end,
the joint distribution $p(v,a)$ and the multiplication of marginal distributions $p(v)p(a)$ 
should be as different as possible. 
Thus, we maximize the mutual information between $v$ and $a$,
which is the KL divergence between $p(v,a)$ and $p(v)p(a)$ as shown in Eq.~\ref{eq:MI}.

Contrastive learning is used to maximize the approximated estimator 
of the KL divergence as in Eq.~\ref{eq:jsd}. 
We can obtain that $I(V, A) \geq \log (S)-L_{S}$ where
\vspace{-3pt}
\begin{equation}
L_{S}=-E_{V}\left[\log \frac{f\left(v_{i}, a_{i}\right)}{\sum_{x_{j} \in V} f\left(v_{j}, a_{j}\right)}\right].
\end{equation}
Here $\left\{v_{j}, a_{i}\right\}(j \neq i)$ is a negative pair and $S$ is the number of all the negative pairs. 
If we set $f=\exp \left(\frac{\left\|v_{i}-c\right\|_{2}^{2}}{\tau}\right)$, 
then $L_{S}$ is equal to the contrastive loss. 
It can be concluded that contrastive loss is a lower bound of mutual information.

For the contrastive learning in our experiment, 
$(v, a)$ are paired visual and semantic features that serve as positive pairs
while $(v^{\prime}, a)$ are negative pairs. 
\section{Conclusion}
\label{sec:conclusion}
In this work, we present a \textbf{S}keleton-based 
\textbf{M}utual \textbf{I}nformation \textbf{E}stimation and maximization framework (SMIE) for
zero-shot action recognition, 
which consists of a global alignment module and a temporal constraint module. 
The global alignment module captures the complex statistical correlations 
between the visual space and the semantic space 
by applying mutual information as the similarity measure. 
The temporal constraint module exploits the inherent temporal information 
to keep the mutual information increasing with the number of observed frames. 
Extensive experiments on three skeleton benchmarks including 
NTU-60, NTU-120, and PKU-MMD datasets verify the effectiveness of our proposed SMIE model. 

\section{Acknowledgements}
\label{sec:acknowledgements}
This work was supported by Shanghai AI Laboratory, 
the National Natural Science Foundation of China No. 61976206 and No. 61832017, 
Beijing Outstanding Young Scientist Program 
NO. BJJWZ\\YJH012019100020098, 
Beijing Academy of Artificial Intelligence (BAAI), 
the Fundamental Research Funds for the Central Universities, 
the Research Funds of Renmin University of China 21XNLG05, 
and Public Computing Cloud, Renmin University of China.

\bibliographystyle{ACM-Reference-Format}
 \balance
 \bibliography{mm2023}


\begin{thebibliography}{47}


\ifx \showCODEN    \undefined \def \showCODEN     #1{\unskip}     \fi
\ifx \showDOI      \undefined \def \showDOI       #1{#1}\fi
\ifx \showISBNx    \undefined \def \showISBNx     #1{\unskip}     \fi
\ifx \showISBNxiii \undefined \def \showISBNxiii  #1{\unskip}     \fi
\ifx \showISSN     \undefined \def \showISSN      #1{\unskip}     \fi
\ifx \showLCCN     \undefined \def \showLCCN      #1{\unskip}     \fi
\ifx \shownote     \undefined \def \shownote      #1{#1}          \fi
\ifx \showarticletitle \undefined \def \showarticletitle #1{#1}   \fi
\ifx \showURL      \undefined \def \showURL       {\relax}        \fi
\providecommand\bibfield[2]{#2}
\providecommand\bibinfo[2]{#2}
\providecommand\natexlab[1]{#1}
\providecommand\showeprint[2][]{arXiv:#2}

\bibitem[Brattoli et~al\mbox{.}(2020)]%
        {Brattoli_2020_CVPR}
\bibfield{author}{\bibinfo{person}{Biagio Brattoli}, \bibinfo{person}{Joseph
  Tighe}, \bibinfo{person}{Fedor Zhdanov}, \bibinfo{person}{Pietro Perona},
  {and} \bibinfo{person}{Krzysztof Chalupka}.} \bibinfo{year}{2020}\natexlab{}.
\newblock \showarticletitle{Rethinking Zero-Shot Video Classification:
  End-to-End Training for Realistic Applications}. In
  \bibinfo{booktitle}{\emph{IEEE Conference on Computer Vision and Pattern
  Recognition (CVPR)}}.
\newblock


\bibitem[Cao et~al\mbox{.}(2017)]%
        {cao2017realtime}
\bibfield{author}{\bibinfo{person}{Zhe Cao}, \bibinfo{person}{Tomas Simon},
  \bibinfo{person}{Shih-En Wei}, {and} \bibinfo{person}{Yaser Sheikh}.}
  \bibinfo{year}{2017}\natexlab{}.
\newblock \showarticletitle{Realtime multi-person 2d pose estimation using part
  affinity fields}. In \bibinfo{booktitle}{\emph{IEEE Conference on Computer
  Vision and Pattern Recognition}}. \bibinfo{pages}{7291--7299}.
\newblock


\bibitem[Carreira and Zisserman(2017)]%
        {carreira2017quo}
\bibfield{author}{\bibinfo{person}{Joao Carreira} {and} \bibinfo{person}{Andrew
  Zisserman}.} \bibinfo{year}{2017}\natexlab{}.
\newblock \showarticletitle{Quo vadis, action recognition? a new model and the
  kinetics dataset}. In \bibinfo{booktitle}{\emph{proceedings of the IEEE
  Conference on Computer Vision and Pattern Recognition}}.
  \bibinfo{pages}{6299--6308}.
\newblock


\bibitem[Cheng et~al\mbox{.}(2020)]%
        {cheng2020skeleton-based}
\bibfield{author}{\bibinfo{person}{Ke Cheng}, \bibinfo{person}{Yifan Zhang},
  \bibinfo{person}{Xiangyu He}, \bibinfo{person}{Weihan Chen},
  \bibinfo{person}{Jian Cheng}, {and} \bibinfo{person}{Hanqing Lu}.}
  \bibinfo{year}{2020}\natexlab{}.
\newblock \showarticletitle{Skeleton-Based Action Recognition With Shift Graph
  Convolutional Network}.
\newblock \bibinfo{journal}{\emph{CVPR}} (\bibinfo{year}{2020}),
  \bibinfo{pages}{180--189}.
\newblock


\bibitem[Du et~al\mbox{.}(2015)]%
        {du2015hierarchical}
\bibfield{author}{\bibinfo{person}{Yong Du}, \bibinfo{person}{Wei Wang}, {and}
  \bibinfo{person}{Liang Wang}.} \bibinfo{year}{2015}\natexlab{}.
\newblock \showarticletitle{Hierarchical recurrent neural network for skeleton
  based action recognition}.
\newblock \bibinfo{journal}{\emph{IEEE Conference on Computer Vision and
  Pattern Recognition}} (\bibinfo{year}{2015}).
\newblock


\bibitem[Frome et~al\mbox{.}(2013)]%
        {frome2013devise}
\bibfield{author}{\bibinfo{person}{Andrea Frome}, \bibinfo{person}{S.~Gregory
  Corrado}, \bibinfo{person}{Jonathon Shlens}, \bibinfo{person}{Samy Bengio},
  \bibinfo{person}{Jeffrey Dean}, \bibinfo{person}{Marc'Aurelio Ranzato}, {and}
  \bibinfo{person}{Tomas Mikolov}.} \bibinfo{year}{2013}\natexlab{}.
\newblock \showarticletitle{DeViSE: A Deep Visual-Semantic Embedding Model}.
\newblock \bibinfo{journal}{\emph{NIPS}} (\bibinfo{year}{2013}).
\newblock


\bibitem[Gan et~al\mbox{.}(2015)]%
        {gan2015exploring}
\bibfield{author}{\bibinfo{person}{Chuang Gan}, \bibinfo{person}{Ming Lin},
  \bibinfo{person}{Yi Yang}, \bibinfo{person}{Yueting Zhuang}, {and}
  \bibinfo{person}{G.~Alexander Hauptmann}.} \bibinfo{year}{2015}\natexlab{}.
\newblock \showarticletitle{Exploring Semantic Inter-Class Relationships (SIR)
  for Zero-Shot Action Recognition}.
\newblock \bibinfo{journal}{\emph{AAAI}} (\bibinfo{year}{2015}),
  \bibinfo{pages}{3769--3775}.
\newblock


\bibitem[Guo et~al\mbox{.}(2022)]%
        {guo2022contrastive}
\bibfield{author}{\bibinfo{person}{Tianyu Guo}, \bibinfo{person}{Hong Liu},
  \bibinfo{person}{Zhan Chen}, \bibinfo{person}{Mengyuan Liu},
  \bibinfo{person}{Tao Wang}, {and} \bibinfo{person}{Runwei Ding}.}
  \bibinfo{year}{2022}\natexlab{}.
\newblock \showarticletitle{Contrastive Learning from Extremely Augmented
  Skeleton Sequences for Self-supervised Action Recognition}. In
  \bibinfo{booktitle}{\emph{Proceedings of the AAAI Conference on Artificial
  Intelligence}}, Vol.~\bibinfo{volume}{36}. \bibinfo{pages}{762--770}.
\newblock


\bibitem[Gupta et~al\mbox{.}(2021)]%
        {gupta2021syntactically}
\bibfield{author}{\bibinfo{person}{Pranay Gupta}, \bibinfo{person}{Divyanshu
  Sharma}, {and} \bibinfo{person}{Kiran~Ravi Sarvadevabhatla}.}
  \bibinfo{year}{2021}\natexlab{}.
\newblock \showarticletitle{Syntactically Guided Generative Embeddings for
  Zero-Shot Skeleton Action Recognition}.
\newblock \bibinfo{journal}{\emph{2021 IEEE International Conference on Image
  Processing (ICIP)}} (\bibinfo{year}{2021}), \bibinfo{pages}{439--443}.
\newblock


\bibitem[Han et~al\mbox{.}(2020)]%
        {han2020learning}
\bibfield{author}{\bibinfo{person}{Zongyan Han}, \bibinfo{person}{Zhenyong Fu},
  {and} \bibinfo{person}{Jian Yang}.} \bibinfo{year}{2020}\natexlab{}.
\newblock \showarticletitle{Learning the redundancy-free features for
  generalized zero-shot object recognition}. In
  \bibinfo{booktitle}{\emph{Proceedings of the IEEE/CVF Conference on Computer
  Vision and Pattern Recognition}}. \bibinfo{pages}{12865--12874}.
\newblock


\bibitem[He et~al\mbox{.}(2016)]%
        {he2016deep}
\bibfield{author}{\bibinfo{person}{Kaiming He}, \bibinfo{person}{Xiangyu
  Zhang}, \bibinfo{person}{Shaoqing Ren}, {and} \bibinfo{person}{Jian Sun}.}
  \bibinfo{year}{2016}\natexlab{}.
\newblock \showarticletitle{Deep residual learning for image recognition}. In
  \bibinfo{booktitle}{\emph{CVPR}}.
\newblock


\bibitem[Hjelm et~al\mbox{.}(2018)]%
        {hjelm2018learning}
\bibfield{author}{\bibinfo{person}{R~Devon Hjelm}, \bibinfo{person}{Alex
  Fedorov}, \bibinfo{person}{Samuel Lavoie-Marchildon}, \bibinfo{person}{Karan
  Grewal}, \bibinfo{person}{Phil Bachman}, \bibinfo{person}{Adam Trischler},
  {and} \bibinfo{person}{Yoshua Bengio}.} \bibinfo{year}{2018}\natexlab{}.
\newblock \showarticletitle{Learning deep representations by mutual information
  estimation and maximization}.
\newblock \bibinfo{journal}{\emph{arXiv preprint arXiv:1808.06670}}
  (\bibinfo{year}{2018}).
\newblock


\bibitem[Hua et~al\mbox{.}(2022)]%
        {hua2022weakly}
\bibfield{author}{\bibinfo{person}{Guoliang Hua}, \bibinfo{person}{Hong Liu},
  \bibinfo{person}{Wenhao Li}, \bibinfo{person}{Qian Zhang},
  \bibinfo{person}{Runwei Ding}, {and} \bibinfo{person}{Xin Xu}.}
  \bibinfo{year}{2022}\natexlab{}.
\newblock \showarticletitle{Weakly-supervised 3D Human Pose Estimation with
  Cross-view U-shaped Graph Convolutional Network}.
\newblock \bibinfo{journal}{\emph{IEEE Transactions on Multimedia}}
  (\bibinfo{year}{2022}).
\newblock


\bibitem[Jasani and Mazagonwalla(2019)]%
        {jasani2019skeleton}
\bibfield{author}{\bibinfo{person}{Bhavan Jasani} {and}
  \bibinfo{person}{Afshaan Mazagonwalla}.} \bibinfo{year}{2019}\natexlab{}.
\newblock \showarticletitle{Skeleton based zero shot action recognition in
  joint pose-language semantic space}.
\newblock \bibinfo{journal}{\emph{arXiv preprint arXiv:1911.11344}}
  (\bibinfo{year}{2019}).
\newblock


\bibitem[Ke et~al\mbox{.}(2017)]%
        {ke2017a}
\bibfield{author}{\bibinfo{person}{Qiuhong Ke}, \bibinfo{person}{Mohammed
  Bennamoun}, \bibinfo{person}{Senjian An}, \bibinfo{person}{Ahmed~Ferdous
  Sohel}, {and} \bibinfo{person}{Farid Boussaïd}.}
  \bibinfo{year}{2017}\natexlab{}.
\newblock \showarticletitle{A New Representation of Skeleton Sequences for 3D
  Action Recognition}.
\newblock \bibinfo{journal}{\emph{CVPR}} (\bibinfo{year}{2017}).
\newblock


\bibitem[Kim and Reiter(2017)]%
        {kim2017interpretable}
\bibfield{author}{\bibinfo{person}{Soo~Tae Kim} {and} \bibinfo{person}{Austin
  Reiter}.} \bibinfo{year}{2017}\natexlab{}.
\newblock \showarticletitle{Interpretable 3D Human Action Analysis with
  Temporal Convolutional Networks}.
\newblock \bibinfo{journal}{\emph{IEEE Computer Society Conference on Computer
  Vision and Pattern Recognition Workshops}} (\bibinfo{year}{2017}),
  \bibinfo{pages}{1623--1631}.
\newblock


\bibitem[Kodirov et~al\mbox{.}(2015)]%
        {7410639}
\bibfield{author}{\bibinfo{person}{Elyor Kodirov}, \bibinfo{person}{Tao Xiang},
  \bibinfo{person}{Zhenyong Fu}, {and} \bibinfo{person}{Shaogang Gong}.}
  \bibinfo{year}{2015}\natexlab{}.
\newblock \showarticletitle{Unsupervised Domain Adaptation for Zero-Shot
  Learning}. In \bibinfo{booktitle}{\emph{2015 IEEE International Conference on
  Computer Vision (ICCV)}}. \bibinfo{pages}{2452--2460}.
\newblock
\urldef\tempurl%
\url{https://doi.org/10.1109/ICCV.2015.282}
\showDOI{\tempurl}


\bibitem[Li et~al\mbox{.}(2019)]%
        {li2019rethinking}
\bibfield{author}{\bibinfo{person}{Kai Li}, \bibinfo{person}{Renqiang~Martin
  Min}, {and} \bibinfo{person}{Yun Fu}.} \bibinfo{year}{2019}\natexlab{}.
\newblock \showarticletitle{Rethinking Zero-Shot Learning - A Conditional
  Visual Classification Perspective}.
\newblock \bibinfo{journal}{\emph{ICCV}} (\bibinfo{year}{2019}).
\newblock


\bibitem[Li et~al\mbox{.}(2021)]%
        {Li_2021_CVPR}
\bibfield{author}{\bibinfo{person}{Linguo Li}, \bibinfo{person}{Minsi Wang},
  \bibinfo{person}{Bingbing Ni}, \bibinfo{person}{Hang Wang},
  \bibinfo{person}{Jiancheng Yang}, {and} \bibinfo{person}{Wenjun Zhang}.}
  \bibinfo{year}{2021}\natexlab{}.
\newblock \showarticletitle{3D Human Action Representation Learning via
  Cross-View Consistency Pursuit}. In \bibinfo{booktitle}{\emph{Proceedings of
  the IEEE/CVF Conference on Computer Vision and Pattern Recognition (CVPR)}}.
  \bibinfo{pages}{4741--4750}.
\newblock


\bibitem[Lin et~al\mbox{.}(2020)]%
        {lin2020ms2l}
\bibfield{author}{\bibinfo{person}{Lilang Lin}, \bibinfo{person}{Sijie Song},
  \bibinfo{person}{Wenhan Yang}, {and} \bibinfo{person}{Jiaying Liu}.}
  \bibinfo{year}{2020}\natexlab{}.
\newblock \showarticletitle{Ms2l: Multi-task self-supervised learning for
  skeleton based action recognition}. In \bibinfo{booktitle}{\emph{Proceedings
  of the 28th ACM International Conference on Multimedia}}.
  \bibinfo{pages}{2490--2498}.
\newblock


\bibitem[Liu et~al\mbox{.}(2019)]%
        {liu2019ntu}
\bibfield{author}{\bibinfo{person}{Jun Liu}, \bibinfo{person}{Amir Shahroudy},
  \bibinfo{person}{Lisboa~Mauricio Perez}, \bibinfo{person}{Gang Wang},
  \bibinfo{person}{Ling-Yu Duan}, {and} \bibinfo{person}{Kot~Alex Chichung}.}
  \bibinfo{year}{2019}\natexlab{}.
\newblock \showarticletitle{NTU RGB+D 120: A Large-Scale Benchmark for 3D Human
  Activity Understanding}.
\newblock \bibinfo{journal}{\emph{IEEE Transactions on Pattern Analysis and
  Machine Intelligence}} (\bibinfo{year}{2019}).
\newblock


\bibitem[Liu et~al\mbox{.}(2020)]%
        {liu2020benchmark}
\bibfield{author}{\bibinfo{person}{Jiaying Liu}, \bibinfo{person}{Sijie Song},
  \bibinfo{person}{Chunhui Liu}, \bibinfo{person}{Yanghao Li}, {and}
  \bibinfo{person}{Yueyu Hu}.} \bibinfo{year}{2020}\natexlab{}.
\newblock \showarticletitle{A benchmark dataset and comparison study for
  multi-modal human action analytics}.
\newblock \bibinfo{journal}{\emph{ACM Transactions on Multimedia Computing,
  Communications, and Applications (TOMM)}} \bibinfo{volume}{16},
  \bibinfo{number}{2} (\bibinfo{year}{2020}), \bibinfo{pages}{1--24}.
\newblock


\bibitem[Mikolov et~al\mbox{.}(2013)]%
        {mikolov2013distributed}
\bibfield{author}{\bibinfo{person}{Tomas Mikolov}, \bibinfo{person}{Ilya
  Sutskever}, \bibinfo{person}{Kai Chen}, \bibinfo{person}{Greg Corrado}, {and}
  \bibinfo{person}{Jeffrey Dean}.} \bibinfo{year}{2013}\natexlab{}.
\newblock \showarticletitle{Distributed Representations of Words and Phrases
  and their Compositionality}.
\newblock \bibinfo{journal}{\emph{Neural Information Processing Systems}}
  (\bibinfo{year}{2013}).
\newblock


\bibitem[Nowozin et~al\mbox{.}(2016)]%
        {nowozin2016f-gan}
\bibfield{author}{\bibinfo{person}{Sebastian Nowozin}, \bibinfo{person}{Botond
  Cseke}, {and} \bibinfo{person}{Ryota Tomioka}.}
  \bibinfo{year}{2016}\natexlab{}.
\newblock \showarticletitle{f-GAN: Training Generative Neural Samplers using
  Variational Divergence Minimization}.
\newblock \bibinfo{journal}{\emph{NIPS}} (\bibinfo{year}{2016}).
\newblock


\bibitem[Qiang et~al\mbox{.}(2021)]%
        {qiang2021robust}
\bibfield{author}{\bibinfo{person}{Wenwen Qiang}, \bibinfo{person}{Jiangmeng
  Li}, \bibinfo{person}{Changwen Zheng}, \bibinfo{person}{Bing Su}, {and}
  \bibinfo{person}{Hui Xiong}.} \bibinfo{year}{2021}\natexlab{}.
\newblock \showarticletitle{Robust local preserving and global aligning network
  for adversarial domain adaptation}.
\newblock \bibinfo{journal}{\emph{IEEE Transactions on Knowledge and Data
  Engineering}} (\bibinfo{year}{2021}).
\newblock


\bibitem[Qiang et~al\mbox{.}(2022)]%
        {qiang2022interventional}
\bibfield{author}{\bibinfo{person}{Wenwen Qiang}, \bibinfo{person}{Jiangmeng
  Li}, \bibinfo{person}{Changwen Zheng}, \bibinfo{person}{Bing Su}, {and}
  \bibinfo{person}{Hui Xiong}.} \bibinfo{year}{2022}\natexlab{}.
\newblock \showarticletitle{Interventional contrastive learning with meta
  semantic regularizer}. In \bibinfo{booktitle}{\emph{International Conference
  on Machine Learning}}. PMLR, \bibinfo{pages}{18018--18030}.
\newblock


\bibitem[Radford et~al\mbox{.}(2021)]%
        {radford2021learning}
\bibfield{author}{\bibinfo{person}{Alec Radford}, \bibinfo{person}{Jong~Wook
  Kim}, \bibinfo{person}{Chris Hallacy}, \bibinfo{person}{Aditya Ramesh},
  \bibinfo{person}{Gabriel Goh}, \bibinfo{person}{Sandhini Agarwal},
  \bibinfo{person}{Girish Sastry}, \bibinfo{person}{Amanda Askell},
  \bibinfo{person}{Pamela Mishkin}, \bibinfo{person}{Jack Clark},
  {et~al\mbox{.}}} \bibinfo{year}{2021}\natexlab{}.
\newblock \showarticletitle{Learning transferable visual models from natural
  language supervision}. In \bibinfo{booktitle}{\emph{International conference
  on machine learning}}. PMLR, \bibinfo{pages}{8748--8763}.
\newblock


\bibitem[Reimers and Gurevych(2019)]%
        {reimers2019sentence-bert}
\bibfield{author}{\bibinfo{person}{Nils Reimers} {and} \bibinfo{person}{Iryna
  Gurevych}.} \bibinfo{year}{2019}\natexlab{}.
\newblock \showarticletitle{Sentence-BERT: Sentence Embeddings using Siamese
  BERT-Networks}.
\newblock \bibinfo{journal}{\emph{EMNLP/IJCNLP}} (\bibinfo{year}{2019}).
\newblock


\bibitem[Schönfeld et~al\mbox{.}(2019)]%
        {schonfeld2019generalized}
\bibfield{author}{\bibinfo{person}{Edgar Schönfeld}, \bibinfo{person}{Sayna
  Ebrahimi}, \bibinfo{person}{Samarth Sinha}, \bibinfo{person}{Trevor Darrell},
  {and} \bibinfo{person}{Zeynep Akata}.} \bibinfo{year}{2019}\natexlab{}.
\newblock \showarticletitle{Generalized Zero- and Few-Shot Learning via Aligned
  Variational Autoencoders}.
\newblock \bibinfo{journal}{\emph{CVPR}} (\bibinfo{year}{2019}).
\newblock


\bibitem[Shahroudy et~al\mbox{.}(2016)]%
        {shahroudy2016ntu}
\bibfield{author}{\bibinfo{person}{Amir Shahroudy}, \bibinfo{person}{Jun Liu},
  \bibinfo{person}{Tian-Tsong Ng}, {and} \bibinfo{person}{Gang Wang}.}
  \bibinfo{year}{2016}\natexlab{}.
\newblock \showarticletitle{NTU RGB+D: A Large Scale Dataset for 3D Human
  Activity Analysis}.
\newblock \bibinfo{journal}{\emph{CVPR}} (\bibinfo{year}{2016}).
\newblock


\bibitem[Shi et~al\mbox{.}(2019)]%
        {shi2019two}
\bibfield{author}{\bibinfo{person}{Lei Shi}, \bibinfo{person}{Yifan Zhang},
  \bibinfo{person}{Jian Cheng}, {and} \bibinfo{person}{Hanqing Lu}.}
  \bibinfo{year}{2019}\natexlab{}.
\newblock \showarticletitle{Two-stream adaptive graph convolutional networks
  for skeleton-based action recognition}. In
  \bibinfo{booktitle}{\emph{Proceedings of the IEEE/CVF Conference on Computer
  Vision and Pattern Recognition}}. \bibinfo{pages}{12026--12035}.
\newblock


\bibitem[Shotton et~al\mbox{.}(2011)]%
        {shotton2011real}
\bibfield{author}{\bibinfo{person}{Jamie Shotton}, \bibinfo{person}{Andrew
  Fitzgibbon}, \bibinfo{person}{Mat Cook}, \bibinfo{person}{Toby Sharp},
  \bibinfo{person}{Mark Finocchio}, \bibinfo{person}{Richard Moore},
  \bibinfo{person}{Alex Kipman}, {and} \bibinfo{person}{Andrew Blake}.}
  \bibinfo{year}{2011}\natexlab{}.
\newblock \showarticletitle{Real-time human pose recognition in parts from
  single depth images}. In \bibinfo{booktitle}{\emph{CVPR 2011}}. Ieee,
  \bibinfo{pages}{1297--1304}.
\newblock


\bibitem[Sylvain et~al\mbox{.}(2020)]%
        {sylvain2020locality}
\bibfield{author}{\bibinfo{person}{Tristan Sylvain}, \bibinfo{person}{Linda
  Petrini}, {and} \bibinfo{person}{Devon Hjelm}.}
  \bibinfo{year}{2020}\natexlab{}.
\newblock \showarticletitle{Locality and Compositionality in Zero-Shot
  Learning}.
\newblock \bibinfo{journal}{\emph{ICLR}} (\bibinfo{year}{2020}).
\newblock


\bibitem[Tang et~al\mbox{.}(2020)]%
        {tang2020zero-shot}
\bibfield{author}{\bibinfo{person}{Chenwei Tang}, \bibinfo{person}{Xue Yang},
  \bibinfo{person}{Jiancheng Lv}, {and} \bibinfo{person}{Zhenan He}.}
  \bibinfo{year}{2020}\natexlab{}.
\newblock \showarticletitle{Zero-shot learning by mutual information estimation
  and maximization}.
\newblock \bibinfo{journal}{\emph{Knowledge-Based Systems}}
  (\bibinfo{year}{2020}).
\newblock


\bibitem[Thoker et~al\mbox{.}(2021)]%
        {thoker2021skeleton}
\bibfield{author}{\bibinfo{person}{Fida~Mohammad Thoker},
  \bibinfo{person}{Hazel Doughty}, {and} \bibinfo{person}{Cees~GM Snoek}.}
  \bibinfo{year}{2021}\natexlab{}.
\newblock \showarticletitle{Skeleton-contrastive 3D action representation
  learning}. In \bibinfo{booktitle}{\emph{Proceedings of the 29th ACM
  international conference on multimedia}}. \bibinfo{pages}{1655--1663}.
\newblock


\bibitem[Tran et~al\mbox{.}(2015)]%
        {tran2015learning}
\bibfield{author}{\bibinfo{person}{Du Tran}, \bibinfo{person}{Lubomir Bourdev},
  \bibinfo{person}{Rob Fergus}, \bibinfo{person}{Lorenzo Torresani}, {and}
  \bibinfo{person}{Manohar Paluri}.} \bibinfo{year}{2015}\natexlab{}.
\newblock \showarticletitle{Learning spatiotemporal features with 3d
  convolutional networks}. In \bibinfo{booktitle}{\emph{Proceedings of the IEEE
  international conference on computer vision}}. \bibinfo{pages}{4489--4497}.
\newblock


\bibitem[Tsai et~al\mbox{.}(2017)]%
        {tsai2017learning}
\bibfield{author}{\bibinfo{person}{Hubert Yao-Hung Tsai},
  \bibinfo{person}{Liang-Kang Huang}, {and} \bibinfo{person}{Ruslan
  Salakhutdinov}.} \bibinfo{year}{2017}\natexlab{}.
\newblock \showarticletitle{Learning Robust Visual-Semantic Embeddings}.
\newblock \bibinfo{journal}{\emph{ICCV}} (\bibinfo{year}{2017}).
\newblock


\bibitem[Tschannen et~al\mbox{.}(2020)]%
        {tschannen2020on}
\bibfield{author}{\bibinfo{person}{Michael Tschannen}, \bibinfo{person}{Josip
  Djolonga}, \bibinfo{person}{K.~Paul Rubenstein}, \bibinfo{person}{Sylvain
  Gelly}, {and} \bibinfo{person}{Mario Lucic}.}
  \bibinfo{year}{2020}\natexlab{}.
\newblock \showarticletitle{On Mutual Information Maximization for
  Representation Learning}.
\newblock \bibinfo{journal}{\emph{ICLR}} (\bibinfo{year}{2020}).
\newblock


\bibitem[Wang et~al\mbox{.}(2023)]%
        {wang2023spatio}
\bibfield{author}{\bibinfo{person}{Jiexin Wang}, \bibinfo{person}{Yujie Zhou},
  \bibinfo{person}{Wenwen Qiang}, \bibinfo{person}{Ying Ba},
  \bibinfo{person}{Bing Su}, {and} \bibinfo{person}{Ji-Rong Wen}.}
  \bibinfo{year}{2023}\natexlab{}.
\newblock \showarticletitle{Spatio-Temporal Branching for Motion Prediction
  using Motion Increments}.
\newblock \bibinfo{journal}{\emph{arXiv preprint arXiv:2308.01097}}
  (\bibinfo{year}{2023}).
\newblock


\bibitem[Wang and Chen(2017)]%
        {wang2017zero-shot}
\bibfield{author}{\bibinfo{person}{Qian Wang} {and} \bibinfo{person}{Ke Chen}.}
  \bibinfo{year}{2017}\natexlab{}.
\newblock \showarticletitle{Zero-Shot Visual Recognition via Bidirectional
  Latent Embedding}.
\newblock \bibinfo{journal}{\emph{International Journal of Computer Vision}}
  (\bibinfo{year}{2017}).
\newblock


\bibitem[Wray et~al\mbox{.}(2019)]%
        {wray2019fine-grained}
\bibfield{author}{\bibinfo{person}{Michael Wray}, \bibinfo{person}{Diane
  Larlus}, \bibinfo{person}{Gabriela Csurka}, {and} \bibinfo{person}{Dima
  Damen}.} \bibinfo{year}{2019}\natexlab{}.
\newblock \showarticletitle{Fine-Grained Action Retrieval Through Multiple
  Parts-of-Speech Embeddings}.
\newblock \bibinfo{journal}{\emph{International Conference on Computer Vision}}
  (\bibinfo{year}{2019}).
\newblock


\bibitem[Xu et~al\mbox{.}(2016)]%
        {xu2016multi-task}
\bibfield{author}{\bibinfo{person}{Xun Xu}, \bibinfo{person}{M.~Timothy
  Hospedales}, {and} \bibinfo{person}{Shaogang Gong}.}
  \bibinfo{year}{2016}\natexlab{}.
\newblock \showarticletitle{Multi-Task Zero-Shot Action Recognition with
  Prioritised Data Augmentation}.
\newblock \bibinfo{journal}{\emph{ECCV}} (\bibinfo{year}{2016}),
  \bibinfo{pages}{343--359}.
\newblock


\bibitem[Yan et~al\mbox{.}(2018)]%
        {yan2018spatial}
\bibfield{author}{\bibinfo{person}{Sijie Yan}, \bibinfo{person}{Yuanjun Xiong},
  {and} \bibinfo{person}{Dahua Lin}.} \bibinfo{year}{2018}\natexlab{}.
\newblock \showarticletitle{Spatial temporal graph convolutional networks for
  skeleton-based action recognition}. In \bibinfo{booktitle}{\emph{AAAI
  Conference on Artificial Intelligence}}, Vol.~\bibinfo{volume}{32}.
\newblock


\bibitem[Zhang et~al\mbox{.}(2020)]%
        {zhang2020semantics}
\bibfield{author}{\bibinfo{person}{Pengfei Zhang}, \bibinfo{person}{Cuiling
  Lan}, \bibinfo{person}{Wenjun Zeng}, \bibinfo{person}{Junliang Xing},
  \bibinfo{person}{Jianru Xue}, {and} \bibinfo{person}{Nanning Zheng}.}
  \bibinfo{year}{2020}\natexlab{}.
\newblock \showarticletitle{Semantics-guided neural networks for efficient
  skeleton-based human action recognition}. In \bibinfo{booktitle}{\emph{IEEE
  Conference on Computer Vision and Pattern Recognition}}.
  \bibinfo{pages}{1112--1121}.
\newblock


\bibitem[Zhang(2012)]%
        {zhang2012microsoft}
\bibfield{author}{\bibinfo{person}{Zhengyou Zhang}.}
  \bibinfo{year}{2012}\natexlab{}.
\newblock \showarticletitle{Microsoft kinect sensor and its effect}.
\newblock \bibinfo{journal}{\emph{IEEE multimedia}} \bibinfo{volume}{19},
  \bibinfo{number}{2} (\bibinfo{year}{2012}), \bibinfo{pages}{4--10}.
\newblock


\bibitem[Zhou et~al\mbox{.}(2023)]%
        {zhou2023self}
\bibfield{author}{\bibinfo{person}{Yujie Zhou}, \bibinfo{person}{Haodong Duan},
  \bibinfo{person}{Anyi Rao}, \bibinfo{person}{Bing Su}, {and}
  \bibinfo{person}{Jiaqi Wang}.} \bibinfo{year}{2023}\natexlab{}.
\newblock \showarticletitle{Self-supervised Action Representation Learning from
  Partial Spatio-Temporal Skeleton Sequences}.
\newblock \bibinfo{journal}{\emph{arXiv preprint arXiv:2302.09018}}
  (\bibinfo{year}{2023}).
\newblock


\bibitem[Zhu et~al\mbox{.}(2018)]%
        {Zhu_2018_CVPR}
\bibfield{author}{\bibinfo{person}{Yi Zhu}, \bibinfo{person}{Yang Long},
  \bibinfo{person}{Yu Guan}, \bibinfo{person}{Shawn Newsam}, {and}
  \bibinfo{person}{Ling Shao}.} \bibinfo{year}{2018}\natexlab{}.
\newblock \showarticletitle{Towards Universal Representation for Unseen Action
  Recognition}. In \bibinfo{booktitle}{\emph{IEEE Conference on Computer Vision
  and Pattern Recognition (CVPR)}}.
\newblock


\end{thebibliography}

\end{document}